\pdfoutput=1

\documentclass[11pt]{article}

\usepackage[]{ACL2023}

\usepackage{times}
\usepackage{latexsym}

\usepackage[T1]{fontenc}

\usepackage[utf8]{inputenc}

\usepackage{microtype}

\usepackage{inconsolata}

\usepackage{multirow}
\usepackage{adjustbox}
\usepackage{array,booktabs}

\usepackage{cleveref}
\usepackage{diagbox}

\title{SamToNe: Improving Contrastive Loss for Dual Encoder Retrieval Models with Same Tower Negatives}

\author{
    Fedor Moiseev\thanks{\quad These authors contributed equally.}
    \quad Gustavo Hern\'{a}ndez \'{A}brego
    \quad Peter Dornbach \\
    \textbf{Imed Zitouni}
    \quad \textbf{Enrique Alfonseca}
    \quad \textbf{Zhe Dong}$^*$\thanks{\quad Corresponding Author.}
    \\
    Google Inc. \\
    \texttt{\small \{femoiseev, gustavoha, dornbach, izitouni, ealfonseca, zhedong\}@google.com}
}

\begin{document}
\maketitle
\begin{abstract}
Dual encoders have been used for retrieval tasks and representation learning with good results. A standard way to train dual encoders is using a contrastive loss with in-batch negatives. In this work, we propose an improved contrastive learning objective by adding queries or documents from the same encoder towers to the negatives, for which we name it as "contrastive loss with SAMe TOwer NEgatives" (SamToNe). By evaluating on question answering retrieval benchmarks from MS MARCO and MultiReQA, and heterogenous zero-shot information retrieval benchmarks (BEIR), we demonstrate that SamToNe can effectively improve the retrieval quality for both symmetric and asymmetric dual encoders. By directly probing the embedding spaces of the two encoding towers via the t-SNE algorithm \citep{JMLR:v9:vandermaaten08a}, we observe that SamToNe ensures the alignment between the embedding spaces from the two encoder towers. Based on the analysis of the embedding distance distributions of the top-$1$ retrieved results, we further explain the efficacy of the method from the perspective of regularisation. 
\end{abstract}

\section{Introduction}

\begin{figure}[!th]
\centering
\includegraphics[width=0.8\columnwidth]{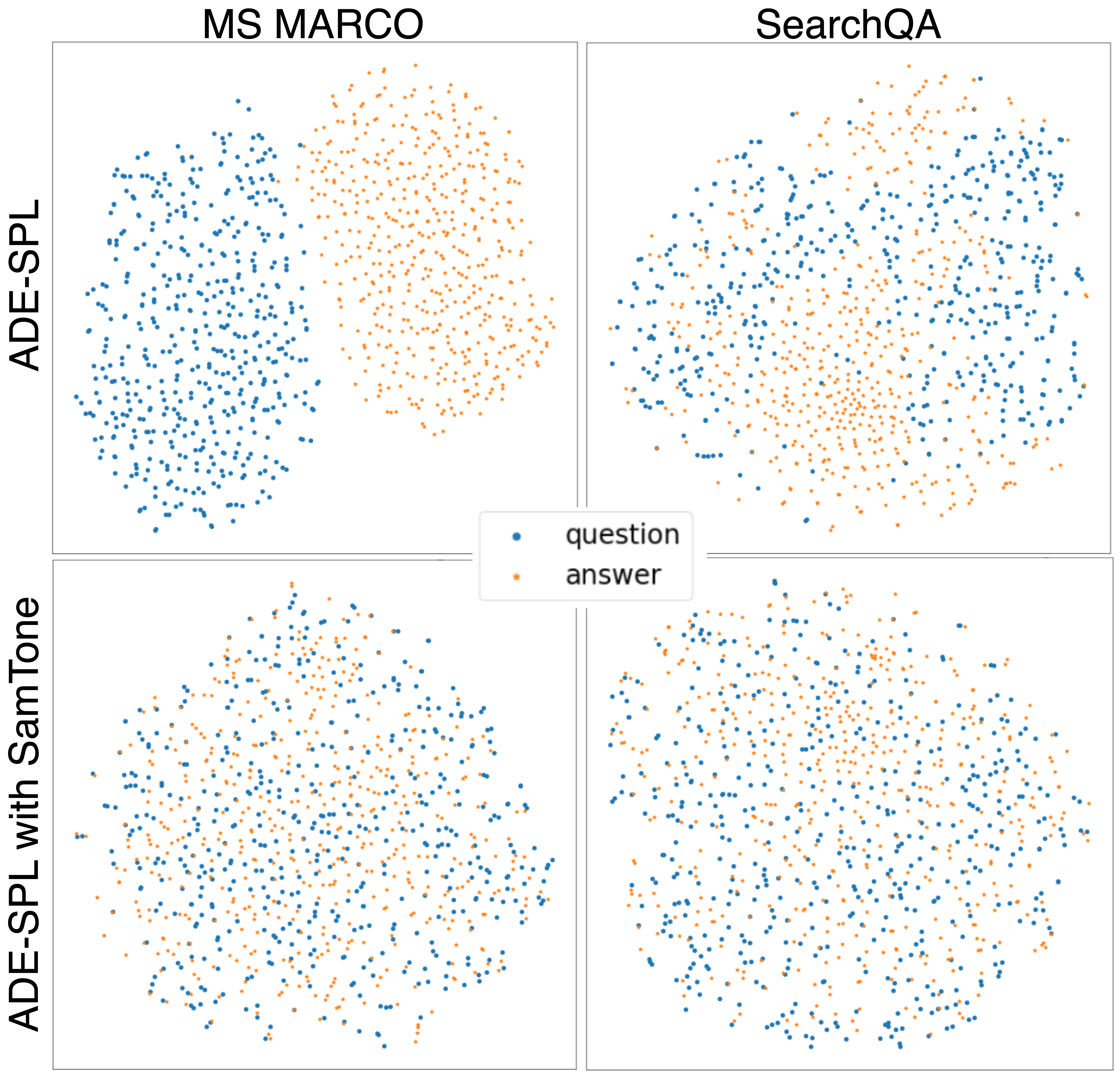}
\caption{\label{fig:embedding_space_analysis}
\footnotesize
Embedding space analyses on MS MARCO and SearchQA show that sharing a projection layer in Asymmetric Dual Encoders (ADE-SPL) \citep{dong-etal-2022-exploring} may not guarantee that the embeddings from the two encoder towers are in a coinciding parameter space. However SamToNe can effectively achieve that. 
}
\vspace{-1.5em}
\end{figure}

The dual encoder architecture applied to information retrieval has shown excellent performance in a wide range of tasks~\citep{Gillick2018EndtoEndRI,karpukhin-etal-2020-dense,GTR,ni-etal-2022-sentence}.

Recently, the Information Retrieval community has transitioned towards Deep Learning models that leverage large unsupervised corpus pre-training~\citep{Devlin2019BERTPO, 2020t5}, which offers more powerful semantic and contextual representation for queries and documents. These models can be successfully applied to scoring tasks, e.g.~\citet{dehghani2017neural}, or retrieval tasks, e.g.~\citet{Gillick2018EndtoEndRI}.
In contrast, classic retrieval models, such as BM25~\citep{bm25}, rely on bag-of-words lexical overlap, term frequency heuristics, inverse document frequency and document length. This type of retrieval models does not require any training and can generalize reasonably well, but they fall short of finding documents that have low term overlap but high semantic similarity.

A dual encoder~\citep{Gillick2018EndtoEndRI,yang-etal-2020-multilingual,karpukhin-etal-2020-dense,reimers-gurevych-2019-sentence} consists of two encoding towers that map queries and documents, respectively, into a shared low-dimensional dense representation, namely, the embedding space. The model is usually optimized by a contrastive loss~\citep{chopra2005learning}, which moves the embeddings of the queries and documents from the same positive examples closer to each other, and the embeddings from negative examples farther away. Training the dual encoder in batches allows to use, for each question, the passages that answer all the other questions within the batch as negatives~\citep{Gillick2018EndtoEndRI}, namely "in-batch negatives". At indexing time, all the documents in a corpus are encoded via bulk inference and indexed. To run retrieval, a query is encoded and its most relevant documents  can be retrieved through Nearest Neighbours Search~\citep{ann1,ann2} over the embedding space using a measure of similarity, e.g. the dot-product or cosine distance of the embedding vectors. 

\paragraph{Motivation.} In this work, we consider two major types of dual encoder architectures: "Symmetric Dual Encoder" (SDE)\footnote{This kind of dual encoders have also been called "Siamese" or "Twin" dual encoders.}, with parameters shared between two encoder towers, and "Asymmetric Dual Encoder" (ADE), with two distinctly parameterized encoder towers. \citet{dong-etal-2022-exploring} demonstrated that sharing projection layers can significantly improve the performance of ADEs. They empirically explained the efficacy of SDE and ADE-SPL by claiming that the shared projection layers help mapping the embeddings of the two encoder towers into a coinciding parameter space. 

By repeating this embedding space analysis on a variety tasks, we find that ADE-SPL may not be enough to ensure that the embedding spaces from two encoder towers are coinciding, as shown in \Cref{fig:embedding_space_analysis}. This motivates us to further improve the dual encoder retrieval quality beyond the architectural change explored in \citet{dong-etal-2022-exploring}. Although the projection layers are shared, our analyses suggest that an extra mechanism, other than using the standard contrastive loss with in-batch negatives, is required to ensure the adjacency of the embeddings of a ground truth pair.

\paragraph{Contributions.} In this paper, we propose an improved training objective for dual encoder models: \textit{contrastive loss with Same Tower Negatives} (\textbf{SamToNe}). In \Cref{section:experiments}, we demonstrate its usefulness on a variety of Information Retrieval tasks, including both tasks with in-task fine-tuning and a zero-shot benchmark suite. Across all the tasks explored, SamToNe performs competitively comparing to the traditional training setup, with a significant improvement on the metrics averaged across tasks. Finally, through an analysis of the produced embeddings, in \Cref{section:analysis}, we further make evident the superiority of SamToNe from the perspective of regularisation.

\section{Method}

\paragraph{Dual Encoder Architecture.} We follow the standard setup of information retrieval: given a query, $q$, and a corpus of retrieval candidates, $\mathcal{P}$, the goal is to retrieve $k$ relevant candidates, $p_k \in \mathcal{P}$. The candidate can be a phrase, a sentence, a passage, or a document.

\begin{figure}[!t]
\centering
\includegraphics[width=0.9\columnwidth]{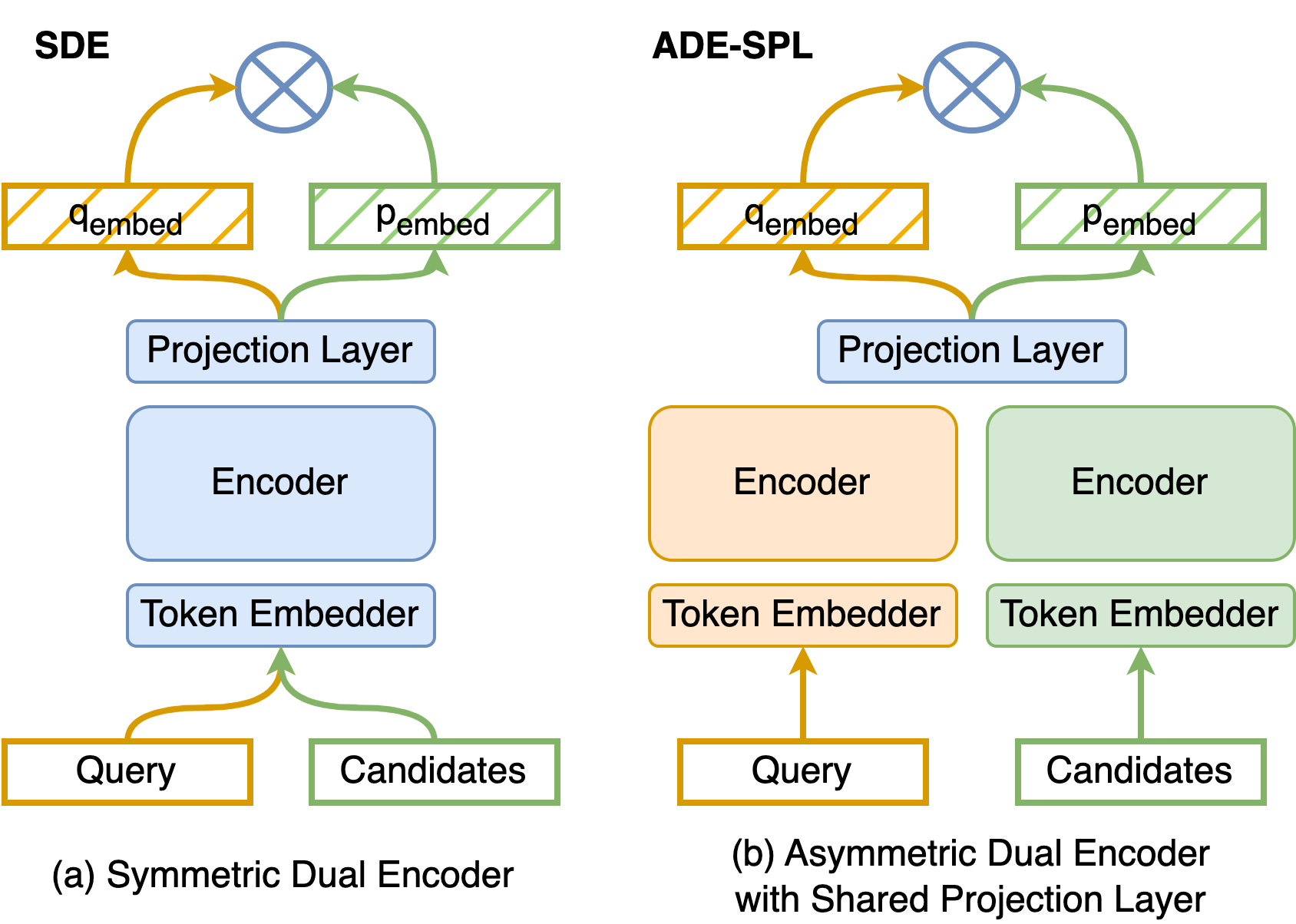}
\caption{\label{fig:dual-encoder-architectures}
\footnotesize
The dual encoder architectures, where the blue components are shared between two encoding paths.}
\vspace{-1em}
\end{figure}

Recent research~\citep{dong-etal-2022-exploring} demonstrated that sharing projection layers can significantly improve the performance of ADEs and we use this shared projection layer for ADEs (ADE-SPL) throughout our experiments. \Cref{fig:dual-encoder-architectures} illustrates the SDE and ADE-SPL architectures we use in this work. Our dual encoders are initialized from pre-trained \texttt{t5.1.1} encoders~\citep{2020t5}. Following~\citet{ni-etal-2022-sentence, dong-etal-2022-exploring}, we encode a query, $q_i$, or a candidate, $p_i$, by averaging the T5 encoder outputs and projecting them to the final embedding vector.

\paragraph{Contrastive Loss.}
A standard way to train a dual encoder model is optimizing an in-batch sampled softmax loss for contrastive learning \citep{Henderson2017EfficientNL}:
\begin{equation}
\footnotesize
  \mathcal{L}_c = \frac{\exp( \mathtt{sim}(q_i, p_i)/ \tau ) } 
  {\sum_{j \in \mathcal{B}}  \exp( \mathtt{sim}(q_i, p_j) / \tau) },
  \label{eq:contrastive-loss}
\end{equation}
where \texttt{sim} is cosine similarity, $\mathcal{B}$ is a mini-batch of examples, and $\tau$ is the softmax temperature. 
$p_i$ is the ground-truth relevant passage for the query $q_i$ in a batch of retrieval candidates $p_*$, where all the other passages $p_k$ ($k\neq i$) are treated as the negative examples for contrastive learning. 

Bi-directional in-batch sampled softmax loss is commonly applied to improve the embedding quality of both towers, where the contrastive loss is computed for both query to passage matching and passage to query matching \citep{bidirect-contr-loss}. We use the bi-directional loss throughout this work.

\paragraph{Same Tower Negatives.}
The in-batch sampled softmax loss is a contrastive loss that only considers the contrastive estimation between the target example pair $\{q_i, p_i\}$, and the in-batch sampled negative pairs $\{q_i,p_j\}$ $(j\neq i)$. 

One way to improve the quality of the retrieval is to improve the contrast among the embeddings of the queries. Therefore, we propose a novel contrastive loss using \textbf{Sam}e \textbf{To}wer \textbf{Ne}gatives, which we abbreviate as \textbf{SamToNe}:
\begin{equation}
\footnotesize
  \mathcal{L}_{S}  = \frac{e^{\mathtt{sim}(q_i, p_i)/ \tau} } 
  {\sum_{j \in \mathcal{B}} e^{\mathtt{sim}(q_i, p_j) / \tau} 
    + \sum_{j \in \mathcal{B}, j \neq i} e^{\mathtt{sim}(q_i, q_j) / \tau} },
  \label{eq:samtone-loss}
\end{equation}
where the second term in the denominator is the contribution from the same tower negatives. 

SamToNe can be interpreted as a regularized version of the in-batch sampled softmax loss, where the term $\sum_{j \in \mathcal{B}, j \neq i} e^{\mathtt{sim}(q_i, q_j) / \tau}$ is a regularizer. When query embeddings are not well distributed, $\max \mathrm{sim}(q_i, q_j) \gg \max \mathrm{sim}(q_i, p_j)$, and the second term in the denominator will dominate the contribution from the negative examples. Thus, it will drive the separation of the query embeddings in contrastive learning. In \Cref{section:analysis}, we provide empirical evidence of the effects of SamToNe as a regularizer of the embedding space.

\citet{ren-etal-2021-pair} proposed an improved contrastive loss, PAIR, which is a hybrid loss 
$\mathcal{L}_{PAIR} = 
- (1-\alpha)\log\mathcal{L}_c 
- \alpha\log\mathcal{L}_P$, 
where 
\begin{equation}
\footnotesize
  \mathcal{L}_{P} = 
  \frac
    {e^{\mathtt{sim}(q_i, p_i)/ \tau}}
    {\sum_{j \in \mathcal{B}, j \neq i}         
     e^{\mathtt{sim}(p_i, p_j) / \tau}}
\end{equation}
penalizes the similarities between passages / documents. 
Despite both SamToNe and PAIR are penalizing the similarities among the same tower inputs, there are two significant differences. 
\textit{Firstly}, SamToNe is hyper-parameter free, while PAIR introduces a new hyper-parameter $\alpha$. This is because SamToNe introduces the new term from an embedding space regularization prospective (see \Cref{section:analysis} for detailed analysis). Therefore SamToNe can be easily applied to both query and document encoders (see \Cref{sec:bidirctional-samtone}), but PAIR needs to introduce yet another hyper-parameter to be applied to both. 
\textit{Secondly}, \citet{ren-etal-2021-pair} mentioned it required a 2-stage training, with the first stage using the PAIR loss, and the second using regular in-batch softmax loss. Due to its self-balancing nature, SamToNe doesn't require multi-stage training. 
A thorough comparison against PAIR can be found in sections \ref{section:experiments} and \ref{section:analysis}.
No added hyper-parameters, single stage training and guaranteed improvement on embedding space quality, make SamToNe much easier to use.

\section{Experiments}
\label{section:experiments}

\begin{figure}[!t]
\centering
    \includegraphics[width=0.468\linewidth]{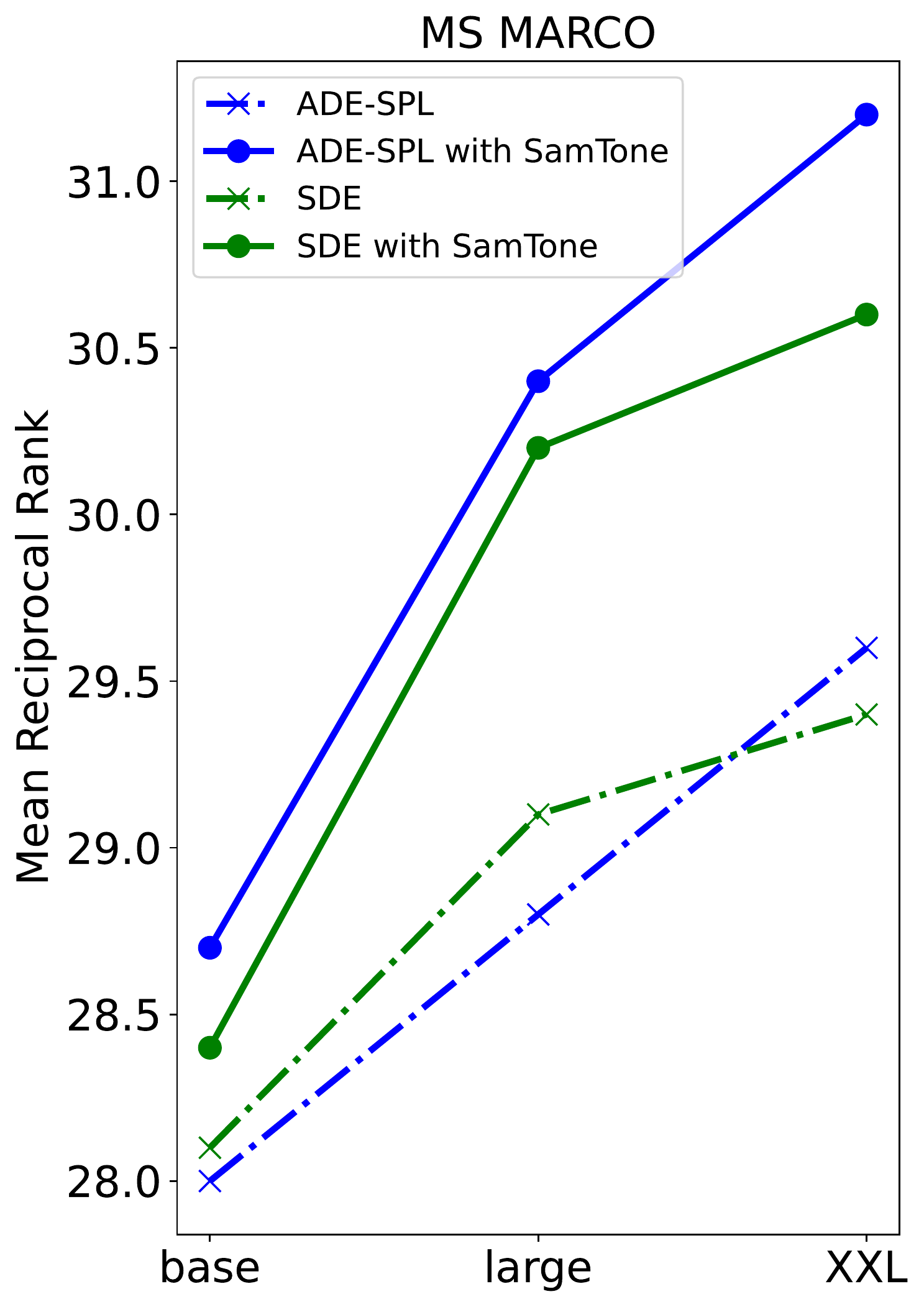}
    \includegraphics[width=0.448\linewidth]{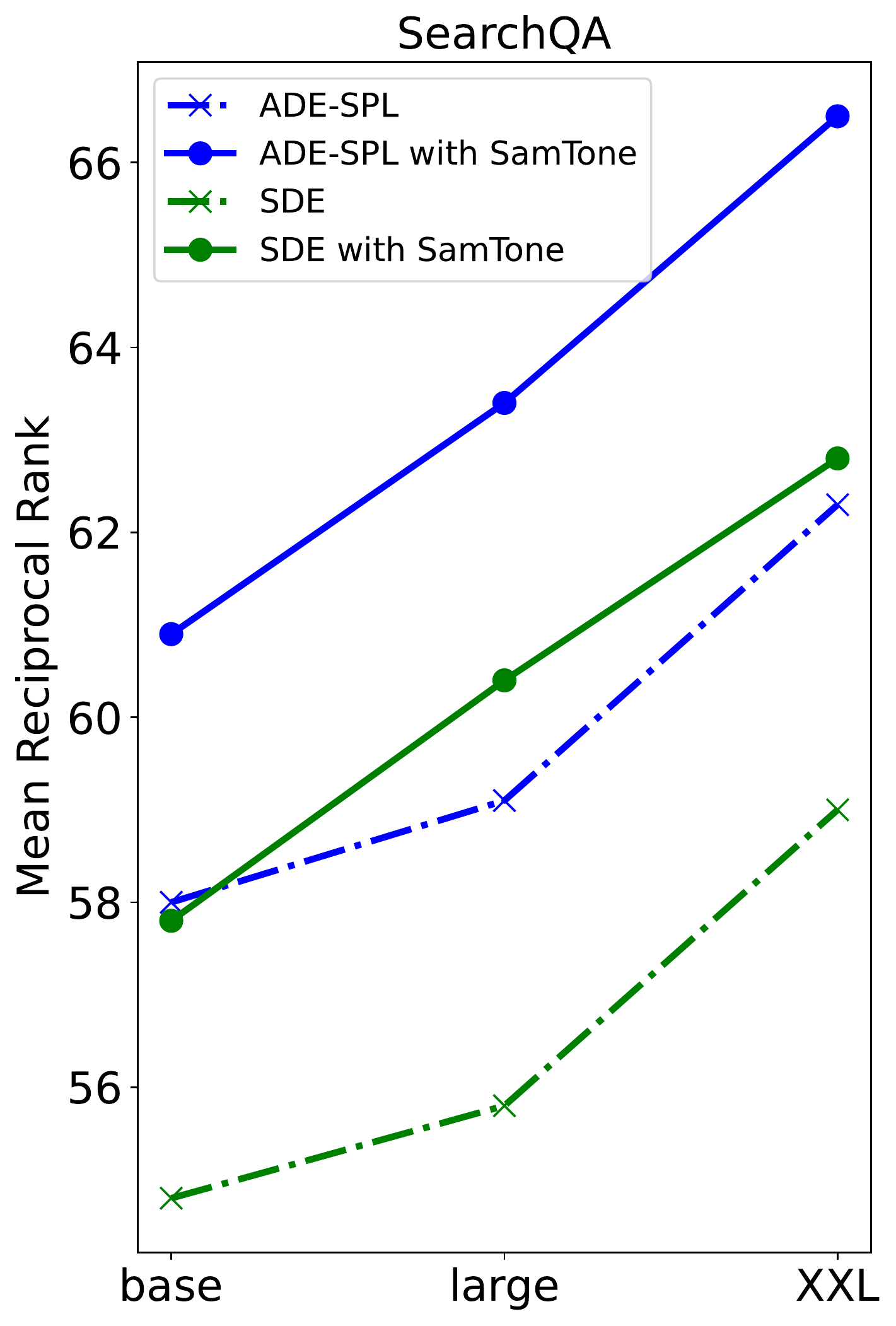}
\caption{\label{figure:mrr-scaling}
\footnotesize
The impact of model sizes on the performance of different dual encoder architectures, measured by MRR on the eval set of MS MARCO (left) and SearchQA (right). 
}
\vspace{-0.7em}
\end{figure}

\begin{figure}[!t]
\centering
\includegraphics[width=0.9\columnwidth]{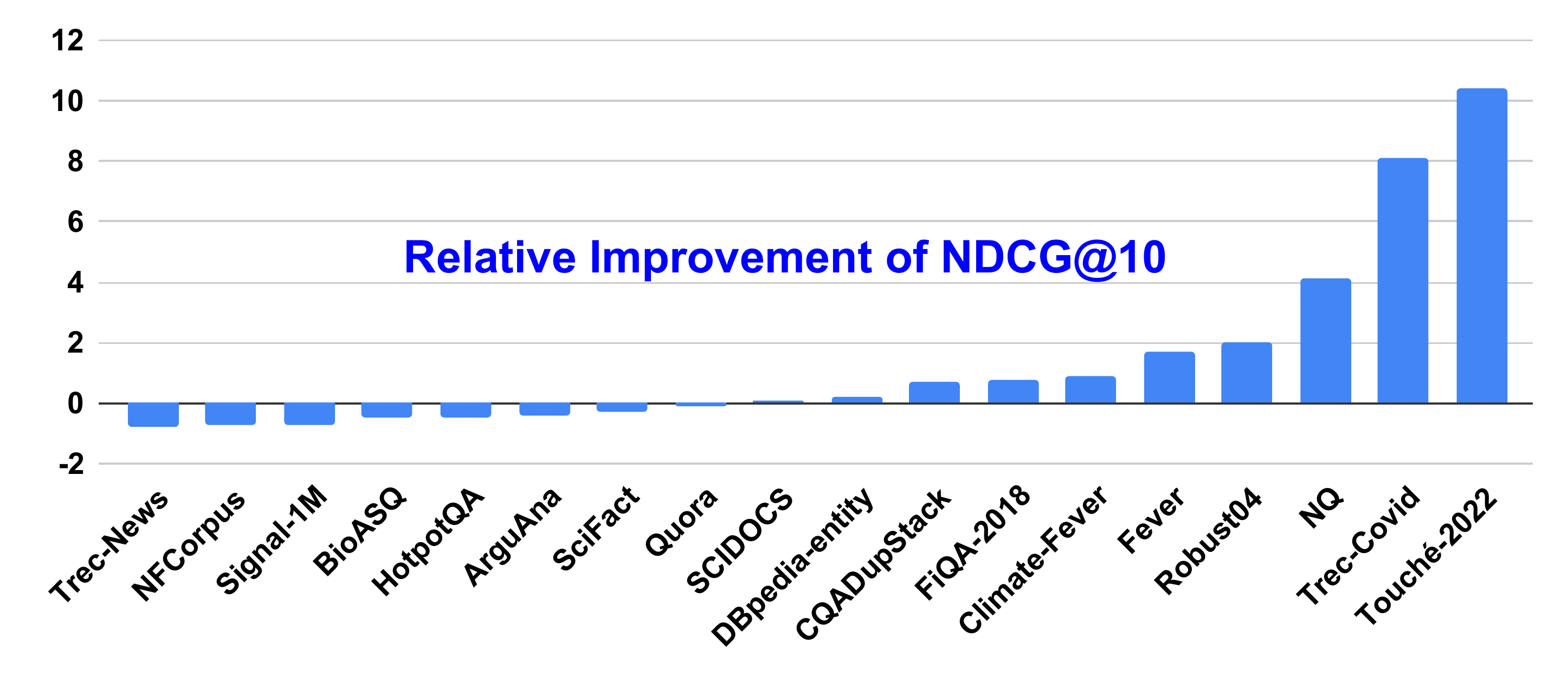}
\vspace{-1em}
\caption{\label{fig:beir-relative-improvements}
\footnotesize
Relative improvement of NDCG@10 ($\%$) on BEIR tasks, by applying SamToNe to SDE. 
}
\vspace{-1em}
\end{figure}

\begin{table*}[!th]
\centering
\adjustbox{max width=\textwidth}{
\begin{tabular}{c|c|cc|cc|cc|cc|cc|cc}
\toprule
    \multirow{2}{*}{\textbf{Model}} & 
    \multirow{2}{*}{\textbf{Loss}} & 
    \multicolumn{2}{c|}{\textbf{MSMARCO}} & 
    \multicolumn{2}{c|}{\textbf{NQ}} &
    \multicolumn{2}{c|}{\textbf{SQuAD}} & 
    \multicolumn{2}{c|}{\textbf{TriviaQA}} & 
    \multicolumn{2}{c|}{\textbf{SearchQA}} & 
    \multicolumn{2}{c}{\textbf{Average}} \\
    & & 
    \textbf{P@1} & \textbf{MRR} & 
    \textbf{P@1} & \textbf{MRR} &
    \textbf{P@1} & \textbf{MRR} & 
    \textbf{P@1} & \textbf{MRR} & 
    \textbf{P@1} & \textbf{MRR} & 
    \textbf{P@1} & \textbf{MRR} \\
\hline \hline
    \multirow{2}{*}{\textbf{ADE}} 
    & Standard\rule{0pt}{2.6ex} & 
    $14.1$ & $26.8$ &
    $53.5$ & $65.2$ & 
    $64.3$ & $74.0$ & 
    $37.9$ & $50.4$ & 
    $41.5$ & $57.2$ & 
    $42.3$ & $54.7$ \\
    & SamToNe & 
    $16.0$ & $28.5$ &
    $52.8$ & $63.9$ & 
    $63.6$ & $73.0$ & 
    $38.4$ & $49.8$ & 
    $\mathbf{49.2}$ & $62.3$ & 
    $44.0$ & $55.5$ \\
    \midrule
    \multirow{3}{*}[-3pt]{\textbf{ADE-SPL}} 
    & Standard &
    $15.7$ & $28.8$ &
    $55.3$ & $67.0$ & 
    $74.5$ & $\mathbf{82.1}$ & 
    $41.7$ & $54.4$ & 
    $42.3$ & $59.1$ &
    $45.9$ & $58.3$ \\
    & SamToNe &
    $\mathbf{17.6}$ & $\mathbf{30.4}$ &
    $\mathbf{55.7}$ & $\mathbf{67.2}$ &
    $73.8$ & $81.7$ &
    $44.0$ & $55.9$ & 
    $48.5$ & $\mathbf{63.4}$ & 
    $\mathbf{47.9}$  & $\mathbf{59.7}$ \\
    \cmidrule{2-14}
    & PAIR & 
    $16.9$ & $29.6$ &
    $55.7$ & $67.0$ & 
    $74.4$ & $82.0$ & 
    $\mathbf{45.0}$ & $\mathbf{56.8}$ & 
    $44.1$ & $60.4$ & 
    $47.2$ & $59.2$ \\
    \midrule
    \multirow{3}{*}[-3pt]{\textbf{SDE}} 
    & Standard &
    $16.1$ & $29.1$ &
    $54.4$ & $66.6$ &
    $74.1$ & $81.9$ &
    $41.4$ & $54.2$ &
    $37.6$ & $55.8$ &
    $44.7$ & $57.5$ \\
    & SamToNe &
    $17.2$ & $30.2$ &
    $54.2$ & $66.4$ &
    $\mathbf{74.6}$ & $\mathbf{82.0}$ &
    $42.1$ & $54.5$ &
    $44.0$ & $60.4$ & 
    $46.4$ & $58.7$ \\
    \cmidrule{2-14}
    & PAIR & 
    $16.1$ & $29.1$ &
    $53.8$ & $66.2$ & 
    $74.13$ & $81.7$ & 
    $41.3$ & $54.5$ & 
    $38.7$ & $56.6$ & 
    $44.7$ & $57.5$ \\
\bottomrule
\end{tabular}
}
\caption{\label{table:qa-retrieval-results} 
\footnotesize
Precision at $1$ (P@1)($\%$) and Mean Reciprocal Rank (MRR)($\%$) on QA retrieval tasks. The best-performing models for each task and metric are highlighted in \textbf{bold}.
}
\vspace{-1em}
\end{table*}

\begin{table}[!t]
\centering
\adjustbox{max width=0.9\textwidth}{
\scriptsize
\begin{tabular}{c|c|c||c|c}
\toprule
\diagbox{Task}{Model} & 
\textbf{SDE} & \textbf{SamToNe} & 
\textbf{BM25} & \textbf{GTR-XXL} \\
\midrule
ArguAna & 
    $\underline{40.2}$ & $39.8$ &
    $31.5$ & $\mathbf{54}$ \\
BioASQ &
    $\underline{40.2}$ & $39.7$ &
    $\mathbf{46.5}$ & $32.4$ \\
Climate-Fever & 
    $31.1$ & $\mathbf{\underline{32}}$ & 
    $21.3$ & $26.7$ \\
CQADupStack & 
    $40.7$ & $\mathbf{\underline{41.4}}$ & 
    $29.9$ & $39.9$ \\
DBpedia-entity & 
    $45.7$ & $\mathbf{\underline{45.9}}$ & 
    $31.3$ & $40.8$ \\
Fever & 
    $68.3$ & $\underline{70}$ & 
    $\mathbf{75.3}$ & $74$ \\
FiQA-2018 & 
    $41.8$ & $\underline{42.6}$ & 
    $23.6$ & $\mathbf{46.7}$ \\
HotpotQA & 
    $\mathbf{\underline{66.9}}$ & 
    $66.4$ & $60.3$ & $59.9$ \\
NFCorpus & 
    $\mathbf{\underline{37.2}}$ & $36.5$ & 
    $32.5$ & $34.2$ \\
NQ & 
    $42.9$ & $\underline{47}$ & 
    $29.9$ & $\mathbf{56.8}$ \\
Quora & 
    $\underline{88.8}$ & $88.7$ & 
    $78.9$ & $\mathbf{89.2}$ \\
Robust04 & 
    $53.5$ & $\mathbf{\underline{55.5}}$ & 
    $40.8$ & $50.6$ \\
SCIDOCS & 
    $22.3$ & $\mathbf{\underline{22.4}}$ & 
    $15.8$ & $15.9$ \\
SciFact & 
    $\mathbf{\underline{68}}$ & $67.7$ & 
    $66.5$ & $66.2$ \\
Signal-1M & 
    $\underline{31.8}$ & $31.1$ & 
    $\mathbf{33}$ & $27.3$ \\
Trec-Covid &
    $53.1$ & $\underline{61.2}$ & 
    $\mathbf{65.6}$ & $50.1$ \\
Trec-News & 
    $\mathbf{\underline{49.2}}$ & $48.4$ &
    $39.8$ & $34.6$ \\
Touché-2022 & 
    $22$ & $\mathbf{\underline{32.4}}$ &
    $36.7$ & $25.6$ \\
\midrule
Average & 
    $46.9$ & $\mathbf{48.3}$ & 
    $42.3$ & $45.8$ \\
\bottomrule
\end{tabular}
}
\caption{\label{table:beir-results} 
\footnotesize
NDCG@10 for zero-shot evaluation on the BEIR benchmark after fine-tuning on MSMarco. The best-performing models for each task are highlighted in \textbf{bold}, while the best scores between \textbf{SDE} and \textbf{SDE~w/~SamToNe} are \underline{underscored}.
}
\vspace{-1em}
\end{table}

\subsection{Question-Answering Retrieval Tasks}
\label{sec:qa-retrieval-tasks}
We evaluate SamToNe on 5 question-answering (QA) retrieval tasks including MS MARCO \citep{bajaj2018msmarco} and MultiReQA \citep{guo-etal-2021-multireqa}. For MS MARCO, the retrieval candidates are relevant passages, and for the 4 tasks in MultiReQA, the retrieval candidates are answer sentences.  

To make a fair comparison across the results of our experiments, the same fine-tuning hyper-parameters are applied to all our model variants. The models are optimized for $20,000$ steps using Adafactor optimizer \citep{shazeer2018adafactor}, with softmax temperature $\tau=0.01$, batch size $512$, and a linearly decaying learning rate starting from $10^{-3}$ to $0$ at the final step. To compare SamToNe and PAIR, we use the hyperparameter $\alpha=0.1$ for PAIR as reported in \citet{ren-etal-2021-pair}, and keep all the other experimental setups identical. SamToNe is applied only on the query side, as it is more robust across different datasets. For experiments and analysis on applying SamToNe on both encoder towers, please refer to \Cref{sec:bidirctional-samtone}. We benchmark the fine-tuned models using precision at $1$ ($P@1$) and mean reciprocal rank (MRR).

As shown in \Cref{table:qa-retrieval-results}, SamToNe greatly improves the retrieval performance of both SDE and ADE-SPL models. Using SamToNe, ADE-SPL models can outperform SDE ones, especially for TriviaQA and SearchQA, by a great margin. Relative to PAIR, SamToNe provides better performance across different datasets in both types of models.

\subsection{Scaling the Model Size}
To assess the impact of the model size, we evaluate the dual encoders 
initialized from \texttt{t5.1.1-base} ($\sim 250$M parameters), \texttt{t5.1.1-large} ($\sim 800$M parameters), and \texttt{t5.1.1-XXL} ($\sim 11$B parameters). \Cref{figure:mrr-scaling} and Appendix \Cref{appendix-table:qa-retrieval-results} show that SamToNe consistently improves the performance of dual encoders across different model sizes. 

\subsection{BEIR Generalization Tasks}
\label{sec:beir}

We further demonstrate the efficacy of the dual encoders trained with SamToNe on BEIR~\cite{thakur2021beir}, a heterogeneous benchmark for zero-shot evaluations.

BEIR has $18$ information retrieval datasets\footnote{MS Marco is excluded from the zero-shot comparison as many baseline models use it as training data.} across $9$ domains, including \textit{Bio-Medical}, \textit{Finance}, \textit{News}, \textit{Twitter}, \textit{Wikipedia}, \textit{StackExchange}, \textit{Quora}, \textit{Scientific}, and \textit{Misc}. The majority of the datasets have binary query relevance labels. The other datasets have 3-level or 5-level relevance judgements. 

As BEIR is evaluating generalization capabilities and SDEs are commonly used for general purpose retrieval \citep{GTR}, we focus on evaluating the impact of SamToNe on BEIR using the SDE architecture. In this evaluation, we reuse the model fine-tuned with MS MARCO, as described in \Cref{sec:qa-retrieval-tasks}. 

Evaluated with the same setting as GTR \citep{GTR}, SamToNe demonstrates strong performance on BEIR, as shown in \Cref{table:beir-results} and \Cref{fig:beir-relative-improvements}. On average, SamToNe improves NDCG@10 by $1.4\%$ for SDE with XXL size. SDE trained with SamToNe significantly outperform BM-25, a sparse retrieval method, and GTR, a dense retrieval method that shares the same architecture and the same model size as SDE but fine-tuned with different corpora.

\subsection{Applying SamToNe to Both Towers}
\label{sec:bidirctional-samtone}
Just as with the query tower, SamToNe can be applied to the document tower which leads to better query-document alignment. However, it is common that the training data contains a large fraction of duplicated documents for a diverse set of queries. For example, only $17\%$ of the documents in the train-split are unique for TriviaQA, but  $98\%$ for MSMARCO. For datasets with a low rate of unique documents, applying SamToNe on the document side will penalize $\mathtt{sim}({p_i, p_j})$ with $p_i = p_j$ and may hinder the performance, as shown in \Cref{table:doc-level-samtone}.

\begin{table}[!t]
\centering
\adjustbox{max width=0.9\textwidth}{
\scriptsize
\begin{tabular}{c|cc|cc}
\toprule
    \multirow{2}{*}{SamToNe} 
    & \multicolumn{2}{c|}{MSMARCO}
    & \multicolumn{2}{c}{TriviaQA}  \\
    & P@1 & MRR & P@1 & MRR \\
\midrule
    W/O SamToNe & $15.7$ & $28.8$ & $41.7$ & $54.4$ \\
    uni-directional & $17.6$ & $30.4$ & $\mathbf{44.0}$ & $\mathbf{55.9}$ \\
    bidirectional & $\mathbf{18.2}$ & $\mathbf{31.0}$ & $41.7$ & $53.3$ \\
\midrule
    $\%$ of unique documents 
    & \multicolumn{2}{c|}{$98\%$} 
    & \multicolumn{2}{c}{$17\%$} \\
\bottomrule
\end{tabular}
}
\caption{
    \label{table:doc-level-samtone} 
    \footnotesize
    Precision at $1$ (P@1)($\%$) and Mean Reciprocal Rank (MRR)($\%$) when comparing ADE-SPL (\texttt{t5.1.1-large} size) trained without SamToNe and with SamToNe applied to the query tower (\textit{uni-directional}) or to both towers (\textit{bidirectional}). The best-performing models for each task and metric are highlighted in \textbf{bold}.
}
\vspace{-1em}
\end{table}

\section{Analysis}
\label{section:analysis}

\subsection{Embedding Space Analysis}

As shown in the top row of \Cref{fig:embedding_space_analysis}, for MS MARCO and SearchQA, ADE-SPL generates two connected but topologically separable embedding spaces. It requires an extra mechanism, beyond the shared projection layers, to ensure the adjacency of the embeddings from a ground truth pair.

SamToNe is proposed as the "force" drawing the embeddings of each ground truth training pair together. Its efficacy is illustrated in the bottom half of \Cref{fig:embedding_space_analysis}.

\subsection{SamToNe: an Embedding Distance Regularizer}
\label{exp:regularizer}

\begin{figure}[!t]
\centering
    \includegraphics[width=0.492\linewidth]{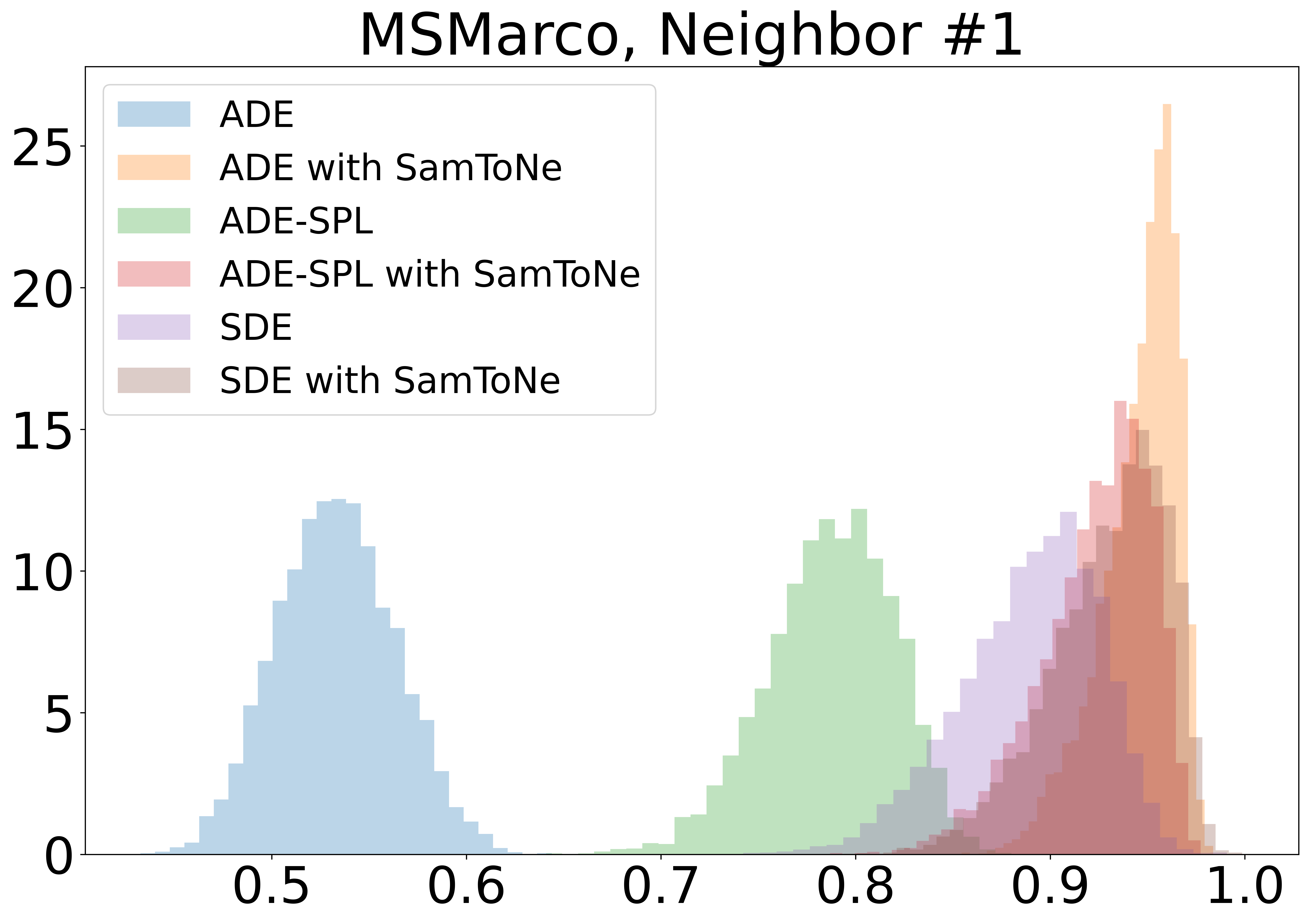}
    \includegraphics[width=0.492\linewidth]{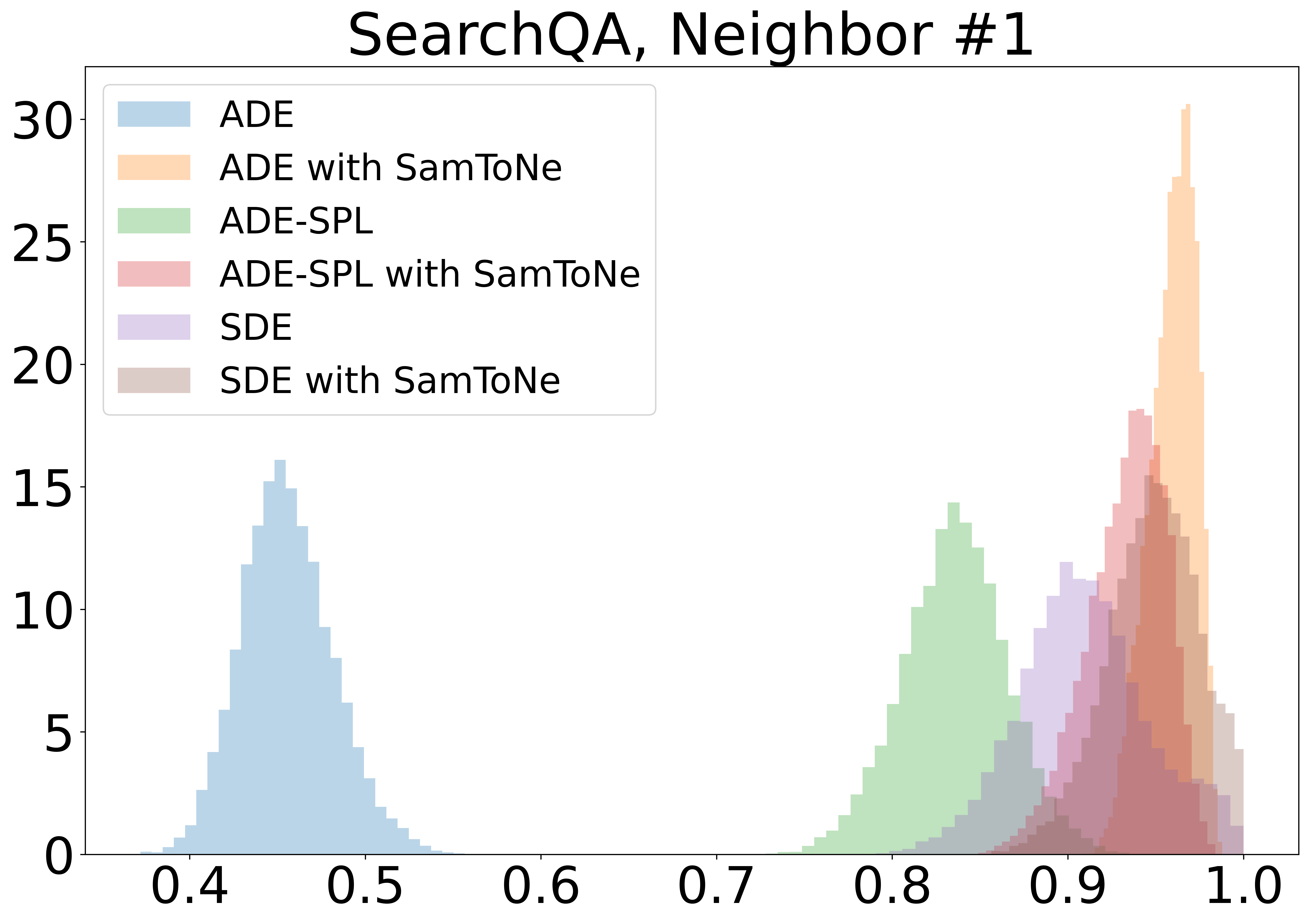}
\caption{\label{fig:embedding_distance_analysis}
\footnotesize
Distributions of cosine similarities between the embeddings of the queries and their \textit{nearest} neighbour documents, for different models trained with or without SamToNe.
}
\vspace{-0.5em}
\end{figure}

To further understand SamToNe's role as a regularizer of embedding distances, we evaluate the distribution of the distances between the embeddings of the queries and their top-$1$ retrieval results in the test set of MS MARCO and SearchQA. The embedding distance is measured by cosine similarity, where $1.0$ means perfect alignment with a range of $[-1.0, 1.0]$. 

As shown in \Cref{fig:embedding_distance_analysis}, SamToNe drastically shifts the distribution of the (query, top-$1$ retrieval result) pairs towards $1.0$, demonstrating the regularizing effect of SamToNe over the embedding distances.

By placing the regularizing query-query similarity terms $e^{\mathtt{sim}(q_i, q_j) / \tau}$ and the standard in-batch negative query-document similarity terms $e^{\mathtt{sim}(q_i, p_j) / \tau}$ together in the denominator with same weight, SamToNe pushes the similarity ratio between query-query and query-documents, $\mathtt{sim}(q_i, q_j)/\mathtt{sim}(q_i, p_j)$, to be centered around $1.0$. This is a \textit{self-balancing} regularization effect. The query and document spaces are set to closely overlap each other and the embeddings of a positive pair are more likely to be located in the same region of the embedding space. 

\begin{figure}[!t]
\centering
\includegraphics[width=0.7\columnwidth]{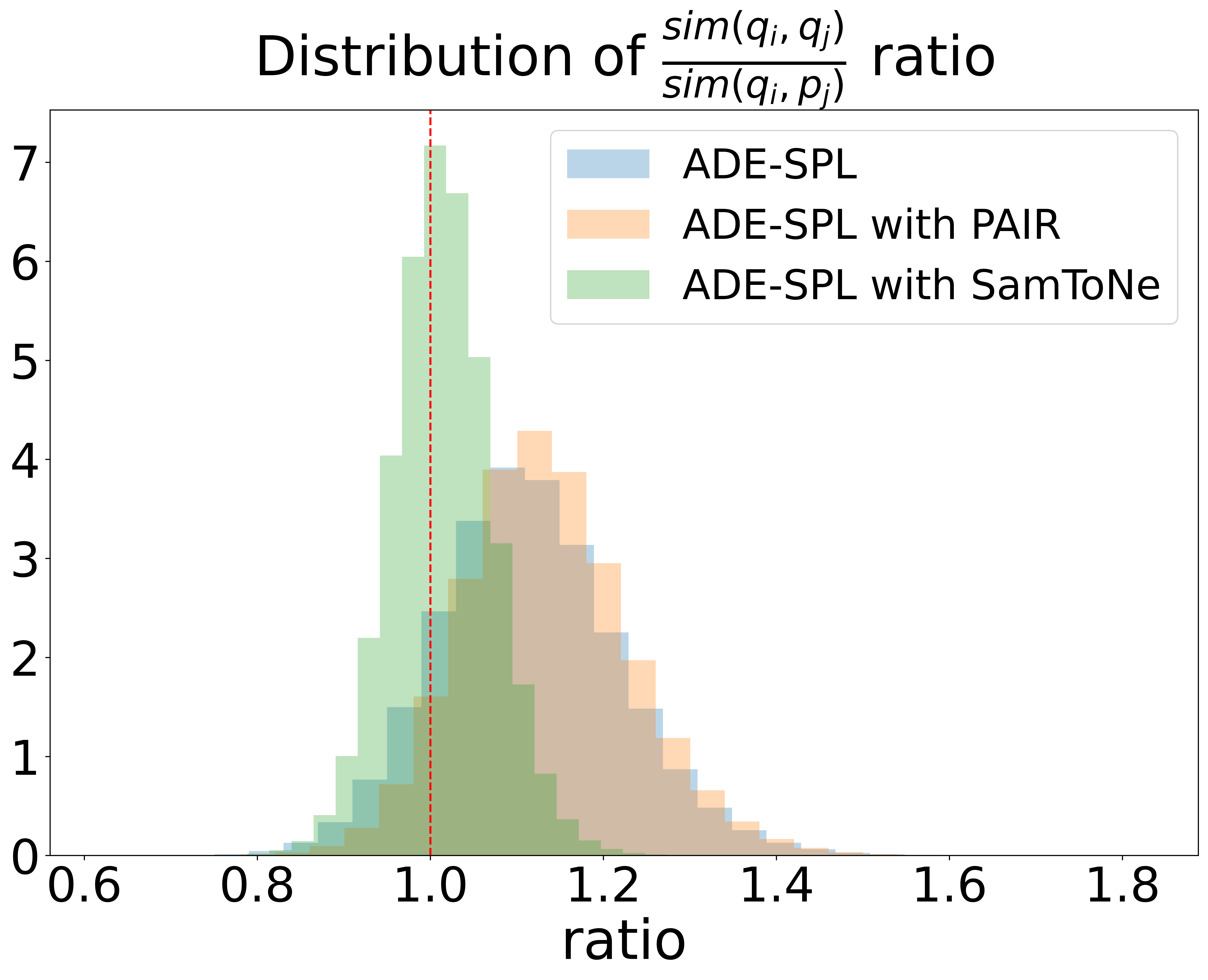}
\caption{\label{fig:qq-qd-ratio}
\footnotesize
Distributions of query-query to query-document similarity ratios for different losses on SearchQA. SamToNe is applied to both query and document sides, and it pushes the ratio to be centered around 1.
}
\vspace{-1.5em}
\end{figure}

To empirically illustrate this effect, we plotted histograms of the $\frac{\mathtt{sim}(q_i, q_j)}{\mathtt{sim}(q_i, p_j)}$ ratios for randomly selected $i$ and $j$ in \Cref{fig:qq-qd-ratio}. The regularization effect only shows when SamToNe is used, but not when PAIR \citep{ren-etal-2021-pair} is. This is because the self-balancing effect does not exist in a hybrid loss such as PAIR.

\section{Conclusions}

Evaluating on QA retrieval tasks and zero-shot generalization benchmarks, we demonstrate that training with SamToNe can significantly improve the dual encoder retrieval quality.  With t-SNE maps of query and document embeddings, we show that the embedding spaces from the two encoding towers of models trained with SamToNe are better aligned. Through the distributions of similarity distances between the embeddings of queries and their nearest neighbours, we empirically explain the efficacy of SamToNe from a regularisation prospective. In general, we recommend using SamToNe to train dual encoders for information retrieval tasks. 

\section{Limitations}

Same tower negatives can be applied to other contrastive losses, e.g. triplet loss \citep{chechik-etal-2010-triplet}. As we are focusing on improving the most popular method to train dual encoder models, i.e. the in-batch sampled softmax loss, we leave the application of same tower negatives to other types of contrastive loss as future work.

While SamToNe has proven to be effective to improve the training of dual encoders, its efficacy may depend on the diversity of the queries used as inputs. In dataset with a large portion of similar queries in the training set, one might need to use masking or other techniques to remove them from the negative computation. Such techniques can also improve the efficacy of SamToNe when applied to both the query and document towers, where SamToNe is currently known to hinder the performance on datasets with a low rate of unique documents, as discussed in \Cref{sec:bidirctional-samtone}.

We leave the in-depth exploration of aforementioned considerations for future works.

\bibliography{acl2023}

\begin{thebibliography}{23}
\expandafter\ifx\csname natexlab\endcsname\relax\def\natexlab#1{#1}\fi

\bibitem[{Bajaj et~al.(2018)Bajaj, Campos, Craswell, Deng, Gao, Liu, Majumder,
  McNamara, Mitra, Nguyen, Rosenberg, Song, Stoica, Tiwary, and
  Wang}]{bajaj2018msmarco}
Payal Bajaj, Daniel Campos, Nick Craswell, Li~Deng, Jianfeng Gao, Xiaodong Liu,
  Rangan Majumder, Andrew McNamara, Bhaskar Mitra, Tri Nguyen, Mir Rosenberg,
  Xia Song, Alina Stoica, Saurabh Tiwary, and Tong Wang. 2018.
\newblock \href {http://arxiv.org/abs/1611.09268} {{MS MARCO}: A human
  generated machine reading comprehension dataset}.

\bibitem[{Chechik et~al.(2010)Chechik, Sharma, Shalit, and
  Bengio}]{chechik-etal-2010-triplet}
Gal Chechik, Varun Sharma, Uri Shalit, and Samy Bengio. 2010.
\newblock \href {http://jmlr.org/papers/v11/chechik10a.html} {Large scale
  online learning of image similarity through ranking}.
\newblock \emph{Journal of Machine Learning Research}, 11(36):1109--1135.

\bibitem[{Chopra et~al.(2005)Chopra, Hadsell, and LeCun}]{chopra2005learning}
Sumit Chopra, Raia Hadsell, and Yann LeCun. 2005.
\newblock Learning a similarity metric discriminatively, with application to
  face verification.
\newblock In \emph{2005 IEEE Computer Society Conference on Computer Vision and
  Pattern Recognition (CVPR'05)}, volume~1, pages 539--546. IEEE.

\bibitem[{Dehghani et~al.(2017)Dehghani, Zamani, Severyn, Kamps, and
  Croft}]{dehghani2017neural}
Mostafa Dehghani, Hamed Zamani, Aliaksei Severyn, Jaap Kamps, and W~Bruce
  Croft. 2017.
\newblock Neural ranking models with weak supervision.
\newblock In \emph{Proceedings of the 40th international ACM SIGIR conference
  on research and development in information retrieval}, pages 65--74.

\bibitem[{Devlin et~al.(2019)Devlin, Chang, Lee, and
  Toutanova}]{Devlin2019BERTPO}
Jacob Devlin, Ming-Wei Chang, Kenton Lee, and Kristina Toutanova. 2019.
\newblock Bert: Pre-training of deep bidirectional transformers for language
  understanding.
\newblock In \emph{NAACL}.

\bibitem[{Dong et~al.(2022)Dong, Ni, Bikel, Alfonseca, Wang, Qu, and
  Zitouni}]{dong-etal-2022-exploring}
Zhe Dong, Jianmo Ni, Daniel~M. Bikel, Enrique Alfonseca, Yuan Wang, Chen Qu,
  and Imed Zitouni. 2022.
\newblock \href {https://aclanthology.org/2022.emnlp-main.640} {Exploring dual
  encoder architectures for question answering}.
\newblock In \emph{Proceedings of the 2022 Conference on Empirical Methods in
  Natural Language Processing}, page 9414–9419. Association for Computational
  Linguistics.

\bibitem[{Gillick et~al.(2018)Gillick, Presta, and
  Tomar}]{Gillick2018EndtoEndRI}
D.~Gillick, A.~Presta, and Gaurav~Singh Tomar. 2018.
\newblock End-to-end retrieval in continuous space.
\newblock \emph{ArXiv}, abs/1811.08008.

\bibitem[{Guo et~al.(2021)Guo, Yang, Cer, Shen, and
  Constant}]{guo-etal-2021-multireqa}
Mandy Guo, Yinfei Yang, Daniel Cer, Qinlan Shen, and Noah Constant. 2021.
\newblock \href {https://aclanthology.org/2021.adaptnlp-1.10} {{M}ulti{R}e{QA}:
  A cross-domain evaluation for{R}etrieval question answering models}.
\newblock In \emph{Proceedings of the Second Workshop on Domain Adaptation for
  NLP}, pages 94--104, Kyiv, Ukraine. Association for Computational
  Linguistics.

\bibitem[{Henderson et~al.(2017)Henderson, Al-Rfou, Strope, Sung, Luk{\'a}cs,
  Guo, Kumar, Miklos, and Kurzweil}]{Henderson2017EfficientNL}
Matthew Henderson, Rami Al-Rfou, B.~Strope, Yun-Hsuan Sung, L{\'a}szl{\'o}
  Luk{\'a}cs, Ruiqi Guo, Sanjiv Kumar, Balint Miklos, and R.~Kurzweil. 2017.
\newblock Efficient natural language response suggestion for smart reply.
\newblock \emph{ArXiv}, abs/1705.00652.

\bibitem[{Johnson et~al.(2021)Johnson, Douze, and Jégou}]{ann2}
Jeff Johnson, Matthijs Douze, and Hervé Jégou. 2021.
\newblock \href {https://doi.org/10.1109/TBDATA.2019.2921572} {Billion-scale
  similarity search with gpus}.
\newblock \emph{IEEE Transactions on Big Data}, 7(3):535--547.

\bibitem[{Karpukhin et~al.(2020)Karpukhin, Oguz, Min, Lewis, Wu, Edunov, Chen,
  and Yih}]{karpukhin-etal-2020-dense}
Vladimir Karpukhin, Barlas Oguz, Sewon Min, Patrick Lewis, Ledell Wu, Sergey
  Edunov, Danqi Chen, and Wen-tau Yih. 2020.
\newblock \href {https://doi.org/10.18653/v1/2020.emnlp-main.550} {Dense
  passage retrieval for open-domain question answering}.
\newblock In \emph{Proceedings of the 2020 Conference on Empirical Methods in
  Natural Language Processing (EMNLP)}, pages 6769--6781, Online. Association
  for Computational Linguistics.

\bibitem[{Ni et~al.(2022)Ni, Hernandez~Abrego, Constant, Ma, Hall, Cer, and
  Yang}]{ni-etal-2022-sentence}
Jianmo Ni, Gustavo Hernandez~Abrego, Noah Constant, Ji~Ma, Keith Hall, Daniel
  Cer, and Yinfei Yang. 2022.
\newblock \href {https://doi.org/10.18653/v1/2022.findings-acl.146}
  {Sentence-t5: Scalable sentence encoders from pre-trained text-to-text
  models}.
\newblock In \emph{Findings of the Association for Computational Linguistics:
  ACL 2022}, pages 1864--1874, Dublin, Ireland. Association for Computational
  Linguistics.

\bibitem[{Ni et~al.(2021)Ni, Qu, Lu, Dai, Ábrego, Ma, Zhao, Luan, Hall, Chang,
  and Yang}]{GTR}
Jianmo Ni, Chen Qu, Jing Lu, Zhuyun Dai, Gustavo~Hernández Ábrego, Ji~Ma,
  Vincent~Y. Zhao, Yi~Luan, Keith~B. Hall, Ming-Wei Chang, and Yinfei Yang.
  2021.
\newblock \href {https://doi.org/10.48550/ARXIV.2112.07899} {Large dual
  encoders are generalizable retrievers}.

\bibitem[{Raffel et~al.(2020)Raffel, Shazeer, Roberts, Lee, Narang, Matena,
  Zhou, Li, and Liu}]{2020t5}
Colin Raffel, Noam~M. Shazeer, Adam Roberts, Katherine Lee, Sharan Narang,
  Michael Matena, Yanqi Zhou, W.~Li, and Peter~J. Liu. 2020.
\newblock Exploring the limits of transfer learning with a unified text-to-text
  transformer.
\newblock \emph{JMLR}, 21/140.

\bibitem[{Reimers and Gurevych(2019)}]{reimers-gurevych-2019-sentence}
Nils Reimers and Iryna Gurevych. 2019.
\newblock \href {https://doi.org/10.18653/v1/D19-1410} {Sentence-{BERT}:
  Sentence embeddings using {S}iamese {BERT}-networks}.
\newblock In \emph{Proceedings of the 2019 Conference on Empirical Methods in
  Natural Language Processing and the 9th International Joint Conference on
  Natural Language Processing (EMNLP-IJCNLP)}, pages 3982--3992, Hong Kong,
  China. Association for Computational Linguistics.

\bibitem[{Ren et~al.(2021)Ren, Lv, Qu, Liu, Zhao, She, Wu, Wang, and
  Wen}]{ren-etal-2021-pair}
Ruiyang Ren, Shangwen Lv, Yingqi Qu, Jing Liu, Wayne~Xin Zhao, QiaoQiao She,
  Hua Wu, Haifeng Wang, and Ji-Rong Wen. 2021.
\newblock \href {https://doi.org/10.18653/v1/2021.findings-acl.191} {{PAIR}:
  Leveraging passage-centric similarity relation for improving dense passage
  retrieval}.
\newblock In \emph{Findings of the Association for Computational Linguistics:
  ACL-IJCNLP 2021}, pages 2173--2183, Online. Association for Computational
  Linguistics.

\bibitem[{Robertson and Zaragoza(2009)}]{bm25}
Stephen Robertson and Hugo Zaragoza. 2009.
\newblock \href {https://doi.org/10.1561/1500000019} {The probabilistic
  relevance framework: Bm25 and beyond}.
\newblock \emph{Found. Trends Inf. Retr.}, 3(4):333–389.

\bibitem[{Shazeer and Stern(2018)}]{shazeer2018adafactor}
Noam Shazeer and Mitchell Stern. 2018.
\newblock \href {https://proceedings.mlr.press/v80/shazeer18a.html} {Adafactor:
  Adaptive learning rates with sublinear memory cost}.
\newblock In \emph{Proceedings of the 35th International Conference on Machine
  Learning}, volume~80 of \emph{Proceedings of Machine Learning Research},
  pages 4596--4604. PMLR.

\bibitem[{Thakur et~al.(2021)Thakur, Reimers, R{\"u}ckl{\'e}, Srivastava, and
  Gurevych}]{thakur2021beir}
Nandan Thakur, Nils Reimers, Andreas R{\"u}ckl{\'e}, Abhishek Srivastava, and
  Iryna Gurevych. 2021.
\newblock \href {https://openreview.net/forum?id=wCu6T5xFjeJ} {{BEIR}: A
  heterogeneous benchmark for zero-shot evaluation of information retrieval
  models}.
\newblock In \emph{Thirty-fifth Conference on Neural Information Processing
  Systems Datasets and Benchmarks Track (Round 2)}.

\bibitem[{van~der Maaten and Hinton(2008)}]{JMLR:v9:vandermaaten08a}
Laurens van~der Maaten and Geoffrey Hinton. 2008.
\newblock \href {http://jmlr.org/papers/v9/vandermaaten08a.html} {Visualizing
  data using t-sne}.
\newblock \emph{Journal of Machine Learning Research}, 9(86):2579--2605.

\bibitem[{Vanderkam et~al.(2013)Vanderkam, Schonberger, Rowley, and
  Kumar}]{ann1}
Dan Vanderkam, Rob Schonberger, Henry Rowley, and Sanjiv Kumar. 2013.
\newblock \href {http://www.google.com/trends/correlate/nnsearch.pdf} {Nearest
  neighbor search in google correlate}.
\newblock Technical report, Google.

\bibitem[{Yang et~al.(2020)Yang, Cer, Ahmad, Guo, Law, Constant,
  Hernandez~Abrego, Yuan, Tar, Sung, Strope, and
  Kurzweil}]{yang-etal-2020-multilingual}
Yinfei Yang, Daniel Cer, Amin Ahmad, Mandy Guo, Jax Law, Noah Constant, Gustavo
  Hernandez~Abrego, Steve Yuan, Chris Tar, Yun-hsuan Sung, Brian Strope, and
  Ray Kurzweil. 2020.
\newblock \href {https://doi.org/10.18653/v1/2020.acl-demos.12} {Multilingual
  universal sentence encoder for semantic retrieval}.
\newblock In \emph{Proceedings of the 58th Annual Meeting of the Association
  for Computational Linguistics: System Demonstrations}, pages 87--94, Online.
  Association for Computational Linguistics.

\bibitem[{Yang et~al.(2019)Yang, Hernandez~Abrego, Yuan, Guo, Shen, Cer, Sung,
  Strope, and Kurzweil}]{bidirect-contr-loss}
Yinfei Yang, Gustavo Hernandez~Abrego, Steve Yuan, Mandy Guo, Qinlan Shen,
  Daniel Cer, Yun-hsuan Sung, Brian Strope, and Ray Kurzweil. 2019.
\newblock \href {https://doi.org/10.24963/ijcai.2019/746} {Improving
  multilingual sentence embedding using bi-directional dual encoder with
  additive margin softmax}.
\newblock In \emph{Proceedings of the Twenty-Eighth International Joint
  Conference on Artificial Intelligence, {IJCAI-19}}, pages 5370--5378.
  International Joint Conferences on Artificial Intelligence Organization.

\end{thebibliography}
\bibliographystyle{acl_natbib}

\appendix

\section{Appendix}
\label{sec:appendix}

\begin{table*}[!t]
\centering
\adjustbox{max width=\textwidth}{
\begin{tabular}{c|c|c|cc|cc|cc|cc|cc|cc}
\toprule
\multirow{2}{*}{\textbf{Model size}} & \multirow{2}{*}{\textbf{Architecture}} & \multirow{2}{*}{\textbf{SamToNe}} & \multicolumn{2}{c}{\textbf{MSMARCO}}        & \multicolumn{2}{c}{\textbf{NQ}}           & \multicolumn{2}{c}{\textbf{SQuAD}}        & \multicolumn{2}{c}{\textbf{TriviaQA}}      & \multicolumn{2}{c}{\textbf{SearchQA}}       & \multicolumn{2}{c}{\textbf{Average}} \\
                                     &                                        &                                   & \textbf{P@1}          & \textbf{MRR}        & \textbf{P@1}        & \textbf{MRR}        & \textbf{P@1}         & \textbf{MRR}       & \textbf{P@1}         & \textbf{MRR}        & \textbf{P@1}         & \textbf{MRR}         & \textbf{P@1}      & \textbf{MRR}     \\
\midrule
\multirow{6}{*}{\textbf{base}}       & \multirow{2}{*}{\textbf{ADE}}          & \textbf{No}                       & $13.8$                & $25.8$              & $48.7$              & $60.1$              & $60.9$               & $70.7$             & $35$                 & $46.3$              & $41.7$               & $57.1$               & $40$              & $52$             \\
                                     &                                        & \textbf{Yes}                      & $15.1$                & $27.1$              & $46.1$              & $57.$               & $59$                 & $68.9$             & $32.5$               & $43.1$              & $45.3$               & $58.5$               & $39.6$            & $50.9$           \\
                                     & \multirow{2}{*}{\textbf{ADE-SPL}}      & \textbf{No}                       & $15.4$                & $28.$               & $50.5$              & $62.1$              & $69.8$               & $78.1$             & $38.8$               & $50.7$              & $41.6$               & $58.$                & $43.2$            & $55.4$           \\
                                     &                                        & \textbf{Yes}                      & $\mathbf{16}$         & $\mathbf{28.7}$     & $\mathbf{50.9}$     & $\mathbf{62.3}$     & $69.9$               & $78.1$             & $\mathbf{40.4}$      & $\mathbf{51.7}$     & $\mathbf{45.8}$      & $\mathbf{60.9}$      & $\mathbf{44.6}$   & $\mathbf{56.3}$  \\
                                     & \multirow{2}{*}{\textbf{SDE}}          & \textbf{No}                       & $15.7$                & $28.1$              & $49.3$              & $61.4$              & $70.2$               & $\mathbf{78.5}$    & $37.7$               & $50.4$              & $36.9$               & $54.8$               & $42$              & $54.6$           \\
                                     &                                        & \textbf{Yes}                      & $15.9$                & $28.4$              & $49.7$              & $61.6$              & $\mathbf{70.4}$      & $0.784$            & $39.4$               & $51.5$              & $41.1$               & $57.8$               & $43.3$            & $55.5$           \\
\midrule
\multirow{6}{*}{\textbf{large}}      & \multirow{2}{*}{\textbf{ADE}}          & \textbf{No}                       & $14.1$                & $26.8$              & $53.5$              & $65.2$              & $64.3$               & $74$               & $37.9$               & $50.4$              & $41.5$               & $57.2$               & $42.3$            & $54.7$           \\
                                     &                                        & \textbf{Yes}                      & $16$                  & $28.5$              & $52.8$              & $63.9$              & $63.6$               & $73$               & $38.4$               & $49.8$              & $49.2$               & $62.3$               & $44$              & $55.5$           \\
                                     & \multirow{2}{*}{\textbf{ADE-SPL}}      & \textbf{No}                       & $15.7$                & $28.8$              & $55.3$              & $67$                & $74.5$               & $\mathbf{82.1}$    & $41.7$               & $54.4$              & $42.3$               & $59.1$               & $45.9$            & $58.3$           \\
                                     &                                        & \textbf{Yes}                      & $\mathbf{17.6}$       & $\mathbf{30.4}$     & $\mathbf{55.7}$     & $\mathbf{67.2}$     & $0.738$              & $0.817$            & $\mathbf{44}$        & $\mathbf{55.9}$     & $\mathbf{48.5}$      & $\mathbf{63.4}$      & $\mathbf{47.9}$   & $\mathbf{59.7}$  \\
                                     & \multirow{2}{*}{\textbf{SDE}}          & \textbf{No}                       & $16.1$                & $29.1$              & $54.4$              & $66.6$              & $74.1$               & $81.9$             & $41.4$               & $54.2$              & $37.6$               & $55.8$               & $44.7$            & $57.5$           \\
                                     &                                        & \textbf{Yes}                      & $17.2$                & $30.2$              & $54.2$              & $66.4$              & $\mathbf{74.6}$      & $82$               & $42.1$               & $54.5$              & $44$                 & $60.4$               & $46.4$            & $58.7$           \\
\midrule
\multirow{6}{*}{\textbf{XXL}}        & \multirow{2}{*}{\textbf{ADE}}          & \textbf{No}                       & $14.9$                & $27.9$              & $57.2$              & $69.2$              & $68.7$               & $77.8$             & $46.1$               & $58.7$              & $47.4$               & $62.7$               & $46.9$            & $59.3$           \\
                                     &                                        & \textbf{Yes}                      & $17$                  & $30$                & $57.5$              & $69$                & $67.7$               & $76.9$             & $47$                 & $58.8$              & $52.7$               & $65.9$               & $48.4$            & $60.1$           \\
                                     & \multirow{2}{*}{\textbf{ADE-SPL}}      & \textbf{No}                       & $16.2$                & $29.6$              & $58.7$              & $70.6$              & $\mathbf{78.3}$      & $\mathbf{85.3}$    & $\mathbf{50.9}$      & $\mathbf{63}$       & $45.7$               & $62.3$               & $50$              & $62.2$           \\
                                     &                                        & \textbf{Yes}                      & $\mathbf{17.7}$       & $\mathbf{31.2}$     & $\mathbf{59.8}$     & $\mathbf{71.4}$     & $77.9$               & $84.8$             & $50.1$               & $61.6$              & $\mathbf{51.9}$      & $\mathbf{66.5}$      & $\mathbf{51.5}$   & $\mathbf{63.1}$  \\
                                     & \multirow{2}{*}{\textbf{SDE}}          & \textbf{No}                       & $15.8$                & $29.4$              & $58.2$              & $70.6$              & $79.2$               & $86$               & $46.9$               & $60.3$              & $40.6$               & $59$                 & $48.1$            & $61.1$           \\
                                     &                                        & \textbf{Yes}                      & $17.1$                & $30.6$              & $58.7$              & $70.8$              & $78.2$               & $85.1$             & $48.3$               & $60.6$              & $46.5$               & $62.8$               & $49.8$            & $62$             \\
\midrule
\multicolumn{3}{c}{Dataset Size (train / test queries / test documents)}                                          & \multicolumn{2}{c}{400776 / 6980 / 8841823} & \multicolumn{2}{c}{106521 / 4131 / 22118} & \multicolumn{2}{c}{87133 / 10485 / 10642} & \multicolumn{2}{c}{335659 / 7776 / 238339} & \multicolumn{2}{c}{629160 / 16476 / 454836} &                   &                 
              
\\
\bottomrule
\end{tabular}
}
\caption{\label{appendix-table:qa-retrieval-results} 
\footnotesize
Precision at $1$(P@1)($\%$) and Mean Reciprocal Rank (MRR)($\%$) on QA retrieval tasks. 
}
\vspace{-1em}
\end{table*}

\end{document}